\pdfoutput=1

\documentclass[11pt]{article}

\usepackage{acl}

\usepackage{times}
\usepackage{latexsym}
\usepackage[style=apa]{biblatex}
\usepackage{pgfplots}
\usepackage{graphicx}
\usepackage{tabularx}
\usepackage{hyperref} 
\usepackage{placeins}

\usepackage[T1]{fontenc}

\usepackage[utf8]{inputenc}

\usepackage{microtype}

\usepackage{graphicx}
\usepackage{caption}

\title{Assessing the Effectiveness of GPT-3 in Detecting False Political Statements: A Case Study on the LIAR Dataset\\}

\author{Mars Gokturk Buchholz \\
Data Scientist and Software Engineer \\
\texttt{mars.gokturk@bugcrowd.com}}

\begin{document}
\maketitle
\begin{abstract}

The detection of political fake statements is crucial for maintaining information integrity and preventing the spread of misinformation in society. Historically, state-of-the-art machine learning models employed various methods for detecting deceptive statements. These methods include the use of metadata (\cite{wang2018liar}), n-grams analysis (\cite{singh2021automatic}), and linguistic (\cite{wu2022feature}) and stylometric (\cite{islam2020using}) features. Recent advancements in large language models, such as GPT-3 (\cite{brown2020language}) have achieved state-of-the-art performance on a wide range of tasks. In this study, we conducted experiments with GPT-3 on the LIAR dataset (\cite{wang2018liar}) and achieved higher accuracy than state-of-the-art models without using any additional meta or linguistic features. Additionally, we experimented with zero-shot learning using a carefully designed prompt and achieved near state-of-the-art performance. An advantage of this approach is that the model provided evidence for its decision, which adds transparency to the model's decision-making and offers a chance for users to verify the validity of the evidence provided.
\end{abstract}

\section{Introduction}

The proliferation of fake news has become a major concern in today's society, with its potential to cause harm to individuals, organizations, and even entire nations. A survey by Pew Research Center in 2016 (\cite{barthel2016americans}) found that 64\% of American adults believe fabricated news stories cause confusion about basic facts of current issues and events. About four-in-ten Americans (39\%) feel confident that they can recognize news that is fabricated, but 23\% say they have shared a made-up news story, with 14\% admitting to sharing a story they knew was fake at the time. 

Organizations, such as PolitiFact\footnote{\url{https://www.politifact.com}}, actively investigate and rate the veracity of comments made by public figures, journalists and organizations. The fact-checking process involves researching the statement, consulting experts, and reviewing relevant documents. Politifact uses a process called the Truth-O-Meter to judge the truth value of statements made by politicians. The process involves fact-checking the statement and assigning a rating on a scale from "True" to "Pants on Fire", which indicates varying degrees of truth. Politifact also provides a detailed explanation of their rating and sources used for each fact-check.  We provide the explanation for each label Politifact uses in Table 1.

\begin{table}[ht]
\begin{tabular}{|p{2cm}|p{4.6cm}|}
\hline
\textbf{Rating} & \textbf{Explanation} \\
\hline
True & The statement is accurate and there is no significant omission of context. \\ 
\hline
Mostly True & The statement is accurate but may require additional context or clarification. \\ 
\hline
Half True & The statement is partially accurate but contains some misleading information or is taken out of context. \\ 
\hline
Mostly False & The statement contains some element of truth but is mostly inaccurate. \\
\hline
False & The statement is inaccurate and contains significant factual errors. \\
\hline
Pants on Fire & The statement is not only inaccurate but also ridiculous or absurd.\\
\hline 
\end{tabular}
\caption{Politifact's Truth-O-Meter labels and descriptions.}
\end{table}

Automated detection of false political statements using natural language processing (NLP) has the potential to streamline the fact-checking process, prevent the spread of false statements on online media and social networks, and increase public awareness of false statements. There has been a growing interest in developing NLP methods to detect and combat false political statements. Historically, some of these methods focused on analyzing and extracting linguistic and stylistic features from statements, as well as augmenting them with specialized knowledge bases or author and context information. However, these approaches face challenges such as intentional language use by bad actors to avoid detection and the potential irrelevance of augmented author information in the long term.

In this paper, we investigate the application of GPT-3 on the task of graded deception detection in political statements. 

\section{Related Work}
\begin{table}[ht]
\begin{tabular}{|p{6.6cm}|}
 \hline
 \textbf{Data Sample}\\
 \hline
 \textbf{id:} 9616.json \\
 \textbf{label:} false \\ 
 \textbf{statement:} “Latinos now make up the majority population in Texas.”\\
 \textbf{subject:} education, legal-issues, population, states \\
\textbf{speaker:} roberto-alonzo \\
\textbf{speaker-job-title:} Attorney \\
\textbf{state-info:} Texas \\
\textbf{party-affiliation}: democrat \\
\textbf{barely-true-c:}	0.0 \\
\textbf{half-true-c:}	1.0 \\
\textbf{false-c:}	0.0 \\
\textbf{mostly-true-c:}	0.0 \\
\textbf{pantsonfire-c:}	0.0 \\
\textbf{context:} a press release\\
\hline
\end{tabular}
\caption{A sample from the training dataset.}
\end{table}
Several studies have investigated automated approaches to detect false statements using the Politifact dataset. These studies have used various features and machine learning techniques to build their models. Features used in these studies include linguistic features such as unigrams, bigrams, and trigrams (\cite{Wang2018};\cite{Rashkin2017}), sentiment scores (\cite{Rashkin2017}; \cite{mubarak2018stance}; \cite{chen2019multi}), named entities (\cite{Rashkin2017};\cite{shu2018fakenewsnet}; \cite{chen2019multi}), and syntax features (\cite{perez2018automatic};\cite{horne2018sampling}). Some studies have also used stylometric features such as readability scores (\cite{mubarak2018stance}) and punctuation features (\cite{Wang2018}).

In this paper, we focus on two studies that utilized the Politifact data to predict the truthfulness of a statement. We will summarize their approaches below.\\\\
\textbf{Lingustic and Stylometric features} One example of using the linguistic and stylometric properties of text to detect false statements is through the use of LIWC.  LIWC stands for "Linguistic Inquiry and Word Count," which is a text analysis software developed by James W. Pennebaker and his colleagues at the University of Texas at Austin. It is designed to analyze the psychological content of natural language and has been widely used in social science research. LIWC features are categories of words that represent specific aspects of language use, such as emotions, social processes, and cognitive processes. Examples of LIWC features include positive emotions like "love," "happy," and "amazing"; negative emotions like "hate," "angry," and "worried"; cognitive processes like "think," "know," and "consider"; and social processes like "friend," "family," and "help". One study that utilized LIWC to detect false statements is "Truth of Varying Shades: Analyzing Language in Fake News and Political Fact-Checking" (\cite{rashkin-etal-2017-truth}). The authors employed the PolitiFact data to make predictions about the truthfulness of political statements. To extract features for prediction, they utilized LIWC. The MaxEnt model, which utilized both text and LIWC features, was found to be the best performing model. These features were instrumental in providing insight into the language used in the statements and the emotional and cognitive tone conveyed, thereby contributing to the prediction. The best model achieved a 6-class prediction macro-averaged F1 score of .22. Using LIWC features produced better performance for some models compared to using only the text feature. However, one of the challenges with using linguistic or stylometric features to detect false statements is that actors may adapt to these detection methods, making it necessary to evaluate the truth value of a statement solely based on the facts.\\\\
\textbf{Meta-data augmentation} The LIAR dataset, constructed by William Yang Wang (\cite{wang2018liar}) using Politifact data, served as the basis for the development of several benchmark models. Both text and meta-information, including the speaker, the speaker's party affiliation, the location of the statement, and its context, were employed as predictive features. The most successful model was a hybrid-CNN model that utilized both text and meta-information, achieving an accuracy of 0.274 on the test dataset. Meta-data played a significant role in prediction, leading to substantial improvements when combined with text. However, using author or context information introduces a challenge, as trends related to the speaker and context may change over time, thereby affecting the model's long-term performance. We will use the same LIAR dataset and benchmarks to evaluate our model's performance.
\section{Dataset}
The experiments were conducted on the LIAR dataset (\cite{wang2018liar}). The dataset contains 12,836 short statements extracted from PolitiFact, a well-known fact-checking website in the United States. These statements were labeled by PolitiFact staff as True, Mostly True, Half True, Barely True, False, or Pants on Fire, according to their truthfulness. The distribution of the six labels is relatively well-balanced, with the exception of the 'Pants on Fire' category, which contains 1,050 instances. The label distribution is shown in Figure 1. The dataset is divided into training, validation, and testing sets using an 80/10/10 split. The average statement length, measured in tokens, is 17.9. The statements primarily date from 2007 to 2016.

\pgfplotsset{width=0.48\textwidth}
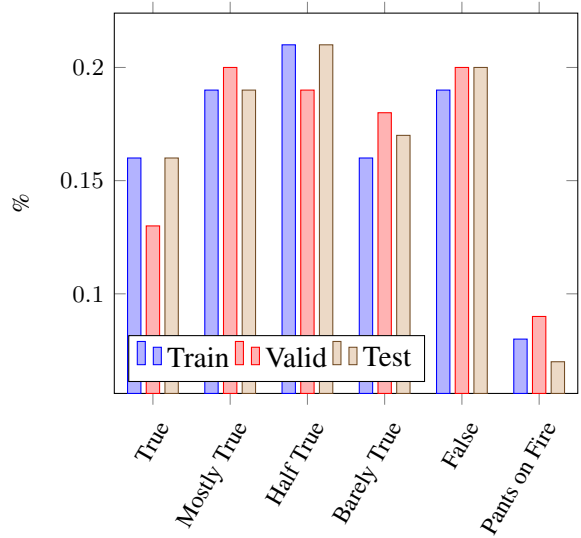
\begin{figure}[htbp]
\begin{tikzpicture}
\begin{axis}[    
ybar,
ylabel style={font=\footnotesize},    ylabel={\%},   
symbolic x coords={true,mostly-true,half-true,barely-true,false,pants-fire},    xtick=data,    ticklabel style={font=\small},    xticklabels={True,Mostly True,Half True,Barely True,False,Pants on Fire},    xticklabel style={rotate=60, anchor=north east},    legend style={at={(0.5,-0.15)},anchor=north,legend columns=-1,
legend pos=south west},
bar width=5pt]]      
\addplot coordinates {(true,0.16) (mostly-true,0.19) (half-true,0.21) (barely-true,0.16) (false,0.19) (pants-fire,0.08)};
\addplot coordinates {(true,0.13) (mostly-true,0.20) (half-true,0.19) (barely-true,0.18) (false,0.20) (pants-fire,0.09)};
\addplot coordinates {(true,0.16) (mostly-true,0.19) (half-true,0.21) (barely-true,0.17) (false,0.20) (pants-fire,0.07)};
\legend{Train,Valid,Test}
\end{axis}
\end{tikzpicture}
\caption{Label distribution in training, validation, and test datasets.} 
\end{figure}

The dataset includes statements covering a wide range of topics, including health care, taxes, education, and the economy. It also contains statements made by politicians from both major political parties in the US. The statements are accompanied by various metadata, such as the speaker's name, party affiliation, and job title, as well as the publication and date of the statement.

\section {Experiments}

We conducted experiments in a single setting, using only the 'statement' column from the LIAR dataset. The models were trained to predict one of the six class labels.

\subsection{Baseline} 
The LIAR dataset (\cite{wang2018liar}) has been widely used for fact-checking research, and several models have been proposed to address the task of detecting fake news. To the best of our knowledge, the best performing model on the LIAR dataset to date is the CNN-hybrid model proposed by Wang (\cite{wang2018liar}), which achieved an accuracy of 0.274 on the test set. 

\subsection{Metric}
In our experiments, we evaluated the performance of our proposed models using the accuracy metric, which is consistent with the LIAR dataset benchmarks.
\begin{equation}
Accuracy = \frac{TP + TN}{TP + TN + FP + FN}
\end{equation} where TP represents true positives, TN represents true negatives, FP represents false positives, and FN represents false negatives.

\subsection{Experiment 1: Fine-tuning GPT-3}
We used the OpenAI Fine-tuning API\footnote{OpenAI Fine-tuning API documentation: \url{https://platform.openai.com/docs/guides/fine-tuning}} to adapt provided GPT-3 base models to the statement classification task.  Our fine-tuning dataset consisted of 10,240 training samples and 1,284 validation samples, which were preprocessed to separate the statement and label columns using predefined tokens.

Considering the rate limits set by OpenAI at the time of the study, we initially fine-tuned the Ada model to identify the optimal hyperparameters, such as the number of epochs and the learning rate multiplier. We experimented with the number of epochs within the range of [2, 3, 4, 5] and the learning rate multiplier within [0.02, 0.04, 0.08, 0.16, 0.2, 0.22]. To account for possible variations in accuracy, each model underwent five training cycles. The optimal hyperparameters were determined to be 5 epochs, a batch size of 256, and a learning rate multiplier of 0.2. We used these parameters to fine-tune the Curie model with the training and validation datasets\footnote{Our datasets, code, and experiment results are publicly available in our GitHub repository: \url{https://github.com/marsgokturk/gpt3-false-political-statement-detection}}.

We utilized the OpenAI fine-tuning API's feature to procure classification metrics on a specified dataset. In our case, we requested this option in the API for our test dataset, which consists of 1,267 samples, to evaluate the model's performance.

The total cost of the fine-tuning process amounted to \$111.27. There were no additional costs associated with serving the fine-tuned model, as the OpenAI API provides classification metrics within the API response for a given dataset. This feature is available if requested as an option in the API request.

Although superior performance might be achievable with different hyperparameters or by using the Davinci model, we opted to fine-tune the Ada model to remain within the rate limits established by OpenAI between March 25th and April 3rd.

\begin{table}
\begin{tabular}{|p{6.6cm}|}
\hline
\textbf{Statement:} "Medicaid spending declined by 1.9 percent in 2012, the second such decline in 47 years." \\
\hline
\textbf{Prompt:} "Act as a research journalist doing fact-checking on the statements given by politicians. You will label each statement as one of "true, mostly-true, half-true, barely-true, false or pants-fire". Use below metric to label each statement:\\
If the statement is accurate and there’s nothing significant missing, then label as "true"\\
If the statement is accurate but needs clarification or additional information, then label as "mostly-true"\\
If the statement is partially accurate but leaves out important details or takes things out of context, then label as "half-true"\\
If the statement contains an element of truth but ignores critical facts that would give a different impression, then label as "barely-true"\\
If the statement is not accurate, then label as "false"\\
If the statement is not accurate and makes a ridiculous claim, then label as "pants-fire"\\
Use only factual data and reliable sources such as major news outlets and fact-checking organizations in the USA. Provide evidence for the labels and the name, date and author of the evidence for your answer.  If you don't have any evidence, say "I don't have any evidence". You answer should be in this format: "Label: <statement label>, Evidence: <evidence>"\\
\hline
\textbf{Output:} "Label = mostly-true, evidence = According to a report by the Kaiser Family Foundation, Medicaid spending did decline by 1.9 percent in 2012, which was the second decline in 47 years. However, the report also notes that the decline was largely due to a decrease in enrollment and utilization, rather than a decrease in spending per enrollee. (source: Kaiser Family Foundation, 'Medicaid Spending and Enrollment Growth: FY 2012 \& 2013, September 2014)" \\
\hline
\end{tabular}
\caption{An example of the model's output for an input statement and the prompt.}
\end{table}
\subsection{Experiment II: Zero-shot learning with prompt-engineering}
In this experiment, we utilized OpenAI Chat Completions API\footnote{OpenAI Chat Completions API documentation: \url{https://platform.openai.com/docs/guides/gpt/chat-completions-api}} with the GPT-3.5-turbo model to perform zero-shot learning through an engineered prompt.

The prompt was meticulously designed to encompass descriptions for each label, adhering to the criteria delineated in Politifact's Truth-O-Meter, and requested factual evidence to substantiate the predicted label. The model was instructed to categorize each statement as "true", "mostly-true", "half-true", "barely-true", "false", or "pants-fire", and to base its label solely on factual information and reliable sources such as major news outlets and fact-checking organizations in the USA.

To assess the model's performance, we employed the same prompt for each test sample, and the accuracy was computed by comparing the gold labels and the extracted output labels. The evidence provided by the model was also documented to gauge its usability and credibility.

We directed the model to generate only one token with a temperature of 0.0. An example of the model's output for an input statement can be observed in Table 3.

This experiment incurred a cost of \$2.00. It is crucial to acknowledge that by creating a more effective prompt, it is conceivable to attain superior accuracy in the zero-shot learning setting.

\section{Results}
\begin{table}
\begin{tabular}{p{4.3cm}|p{2.3cm}}
\textbf{Model} & \textbf{Test Accuracy} \\
\hline
Baseline CNN using only the statement &  0.270 \\
\hline
Baseline CNN-hybrid using the statement+meta-data &  0.274 \\
\hline
Curie fine-tuned model using only the statement &  \textbf{0.295} \\
\hline
Zero-shot setting with the same prompt using only the statement &  0.266 \\
\hline
\end{tabular}
\caption{Comparison of baseline and GPT-3 models.}
\end{table}

Our experiments show that fine-tuned GPT-3 models outperform the best baseline CNN-hybrid model in political statement classification.  Although the zero-shot model's performance is lower, it is still competitive with the baseline model and provides transparency by giving evidence for predictions. Table 4 summarizes the test accuracy results of our experiments.

In the Appendix, we provide a few representative examples of the statement, true label, predicted label, and the model's evidence. These examples give a deeper understanding of the model's performance and reasoning.

\section{Discussion and Conclusion}
\begin{table}
\begin{tabular}{p{2.2cm}|p{4.4cm}}
\textbf{statement} &  "Bernie Sanders voted against the Brady Bill -- background checks and waiting periods."\\
\hline
\textbf{Politifact label} & mostly-true \\
\hline
\textbf{Predicted label;evidence} & false; "bernie sanders voted for the brady bill in 1993, which established background checks and waiting periods for firearm purchases. (source: politifact, "bernie sanders voted for brady bill, but also voted to shield gun makers," february 18, 2016)"   
\end{tabular}
\caption{Disagreement between Politifact label and Predicted label.}
\end{table}

The training data for GPT-3, which only extends up to September 2021, might limit its effectiveness in fact-checking tasks that require up-to-date information. A potential solution for future research could be the use of Retrieval-Augmented Generation (RAG; \cite{DBLP:journals/corr/abs-2005-11401}). This method could provide the model with relevant and current information for a given statement, enabling it to predict the statement's truth value and supply factual evidence to support its prediction. Consequently, this could enhance the model's ability to make accurate predictions, especially when dealing with recent or current events.

In addition, future studies could explore few-shot learning scenarios. By fine-tuning the model with a well-selected set of examples, performance might surpass that of the zero-shot setting. 

The GPT-3 model displayed a complex behavior by not consistently employing or aligning with labels from Politifact data, which suggests that it isn't simply repeating previously absorbed information. Its utilization of Politifact data for label generation sparks questions concerning potential data leaks in contrast to its proficiency for information retrieval during truth-value assessments.  As demonstrated by an example in Table 5, an intriguing area for additional exploration involves instances when models and fact-checkers interpret the same evidence but reach different conclusions. A deeper understanding of these conflicts and devising a robust framework to reconcile these differences could be significantly impactful for future research.

While the fine-tuned GPT-3 model outperformed the state-of-the-art model in classification accuracy, the utility of label-only models for political statement classification remains debatable. The necessity of a six-label classification system, as opposed to a simpler one, may also warrant reconsideration.

\printbibliography
\appendix
\section{Sample Predictions}
\label{sec:appendix}
The table below showcases representative examples of the model's predictions. It presents the statement being evaluated, the true label, the predicted label, and the supporting evidence. 

\begin{table*}
\begin{tabularx}{\textwidth}{|p{3cm}|p{1.3cm}|p{1.5cm}|X|}
\hline
\textbf{Statement} & \textbf{True Label} & \textbf{Predicted Label} & \textbf{Zero-Shot Model's Evidence} \\ 
\hline
Says by rejecting an expansion of Medicaid under the new law, Georgia was depriving 25,000 veterans of health care coverage. & True & False & according to politifact, "there’s no evidence that 25,000 veterans in Georgia would gain health care coverage if the state expanded Medicaid." Additionally, the statement ignores the fact that many veterans are already eligible for health care through the Department of Veterans Affairs. (source: \url{https://www.politifact.com/factchecks/2014/mar/...}) \\ 
\hline
Under Rosemary Lehmberg, the Travis County D.A.'s office convened the grand jury that indicted Rick Perry. & True & False & "Under Lehmberg, the Travis County District Attorney's office had convened a grand jury to investigate allegations that Perry had abused his power by vetoing funding for the state's public integrity unit, which was housed in the Travis County District Attorney's office." - politifact, "Rick Perry says he was right to veto funding for DA's office," August 19, 2014.\\ 
\hline
Foreign Chinese prostitution money is allegedly behind the groups funding Congressman Sean Duffy's Republican Majority. & Pants-Fire & Pants-Fire & I don't have any evidence. \\ 
\hline
Says the unemployment rate for college graduates is 4.4 percent and over 10 percent for noncollege-educated. & True & Mostly-True & according to the Bureau of Labor Statistics, as of August 2021, the unemployment rate for individuals with a bachelor's degree or higher is 3.2 percent, while the unemployment rate for those with less than a high school diploma is 9.0 percent and for those with a high school diploma but no college is 6.0 percent. However, the statement is mostly true as it is close to the actual figures and the difference between the two rates is significant. Source: \url{https://www.bls.gov/web/empsit/cpsee_e16.htm} \\
\hline
Building a wall on the U.S.-Mexico border will take literally years. & True & True & "According to a report by the Government Accountability Office, it took an average of 4.5 years to complete the construction of the 654 miles of fencing already in place along the border as of 2015." (source: NPR, "How long would it take to build Trump's wall?," January 25, 2017)\\
\hline
\end{tabularx}
\end{table*}

\end{document}